\documentclass[letterpaper, 10 pt, conference]{ieeeconf}  % Comment this line out if you need a4paper

\IEEEoverridecommandlockouts                              % This command is only needed if 
\usepackage[utf8]{inputenc}
\usepackage[T1]{fontenc}
\usepackage{microtype}
\usepackage[dvips]{graphicx}
\usepackage{xcolor}
\usepackage{times}
\usepackage{float}
\usepackage{amsmath}
\usepackage{amssymb}
\usepackage{mathabx}
\usepackage{dsfont}
\usepackage[ruled,vlined,noresetcount]{algorithm2e}
\usepackage{cancel}
\usepackage{multirow}
\usepackage{subcaption}
\usepackage{blindtext}
\usepackage{hyperref}
\usepackage{amssymb}
\usepackage{tikz}
\usetikzlibrary{shapes,arrows,calc,positioning}
\usepackage{hyperref}
\usepackage{wrapfig,lipsum,booktabs}

\title{\LARGE \bf
Learning Control from Raw Position Measurements}

\author{Fabio Amadio$^1$, Alberto Dalla Libera$^1$, Daniel Nikovski$^2$,  Ruggero Carli$^1$, Diego Romeres$^2$
    \thanks{$^1$ Fabio Amadio, Alberto Dalla Libera, and Ruggero Carli are with the Department of Information Engineering, University of Padova, Via Gradenigo 6/B, 35131 Padova, Italy [fabio.amadio@phd.unipd.it, dallaliber@dei.unipd.it, carlirug@dei.unipd.it].}
    \thanks{$^3$ Diego Romeres and Daniel Nikovski are with Mitsubishi Electric Research Laboratories (MERL), Cambridge, MA 02139 [romeres@merl.com, nikovski@merl.com]  .}%
}%

\begin{document}

\maketitle
\thispagestyle{empty}
\pagestyle{empty}

%%%%%%%%%%%%%%%%%%%%%%%%%%%%%%%%%%%%%%%%%%%%%%%%%%%%%%%%%%%%%%%%%%%%%%%%%%%%%%%%
\begin{abstract}
We propose a Model-Based Reinforcement Learning (MBRL) algorithm named VF-MC-PILCO, specifically designed for application to mechanical systems where velocities cannot be directly measured. This circumstance, if not adequately considered, can compromise the success of MBRL approaches. To cope with this problem, we define a velocity-free state formulation which consists of the collection of past positions and inputs. Then, VF-MC-PILCO uses Gaussian Process Regression to model the dynamics of the velocity-free state and optimizes the control policy through a particle-based policy gradient approach. We compare VF-MC-PILCO with our previous MBRL algorithm, MC-PILCO4PMS, which handles the lack of direct velocity measurements by modeling the presence of velocity estimators. Results on both simulated (cart-pole and UR5 robot) and real mechanical systems (Furuta pendulum and a ball-and-plate rig) show that the two algorithms achieve similar results. Conveniently, VF-MC-PILCO does not require the design and implementation of state estimators, which can be a challenging and time-consuming activity to be performed by an expert user.
\end{abstract}

%%%%%%%%%%%%%%%%%%%%%%%%%%%%%%%%%%%%%%%%%%%%%%%%%%%%%%%%%%%%%%%%%%%%%%%%%%%%%%%%
\section{Introduction}
Model-Based Reinforcement Learning (MBRL) \cite{polydoros2017survey} proved to be a promising strategy to overcome the challenges of delivering Reinforcement Learning (RL) \cite{sutton2018reinforcement} solutions to real-world problems. In fact, standard Model-Free RL algorithms usually require a massive amount of interaction with the systems to solve the assigned task. This requirement might be unfeasible in many real-world applications, e.g. control of mechanical systems and robotics, due to the limited time available and the risk of damaging the devices involved in such a long training phase. On the other hand, MBRL uses the collected data to train a predictive model of the environment and updates the policy based on model simulations. With this strategy, we are able to extract more valuable information from the available data and increase data efficiency \cite{atkeson1997comparison}.

Nevertheless, the effectiveness of MBRL methods strongly depends on how accurately the trained model can simulate the behavior of the environment. For this reason, it is necessary to adopt stochastic models in order to capture uncertainty about predictions. Different classes of models have been employed, from Gaussian Processes (GPs) \cite{williams2006gaussian} in \cite{deisenroth2011pilco,deisenroth2011learning, berkenkamp2017safe, chatzilygeroudis2017black}, to ensembles of probabilistic deep neural networks in \cite{chua2018deep,kurutach2018model}. 

The application of this kind of method to real-world environments is affected by another major problem: the full state of a real system is often only partially measurable. For instance, when dealing with mechanical systems, joint positions can be measured by means of proper sensors, e.g. encoders, while velocities can only be estimated from the history of sampled positions. In our previous work \cite{amadio2021tro}, we proposed an MBRL algorithm, called MC-PILCO4PMS, specifically tailored to deal with Partially Measurable Systems and take correctly into account the presence of online and offline state observers. It proved able to robustly learn from scratch how to control mechanical systems, in both simulated and real environments even when the velocity is not directly measurable. However, the tuning of accurate filters and state estimators could be particularly challenging and time-consuming for systems affected by significant noise. 

In this work, we present an alternative approach, called VF-MC-PILCO, that completely circumvents the necessity of estimating velocities, by working only with the history of measured positions and applied control actions. We adopted a Velocity-Free (VF) predictive model, similar to the one proposed in \cite{DFikin,DFRL}, together with a control policy whose input depends only on positions. Compared to the works in \cite{DFikin,DFRL}, which were focused on the modeling part, in this work we propose a complete VF solution to the RL problem. The proposed method is first tested in two simulated systems with an increasing number of DoF i.e., a cart-pole and a 6 DoF UR5 robot.
Then, VF-MC-PILCO is tested in two real systems with an increasing level of difficulty for the velocity estimation i.e., a Furuta pendulum equipped with encoders and a ball-and-plate system equipped with an external camera to infer the ball positions. VF-MC-PILCO correctly solved all the tasks with performance similar to the one obtained by MC-PILCO4PMS.
%Remarkably, VF-MC-PILCO managed to work directly with raw measurements, despite the presence of significant noise.
To solve such tasks, MC-PILCO4PMS must accurately reproduce the online filter employed inside the policy optimization algorithm and implement an offline filtering procedure for estimating the velocities needed for modeling.
On the contrary, VF-MC-PILCO, by working directly with raw measurements, presents an alternative way that requires less effort and expertise from the user, obtaining similar performance despite the presence of significant noise. The comparisons are carried against MC-PILCO4PMS because this algorithm was shown to outperform other s.o.t.a. MBRL algorithms in \cite{amadio2021tro}.

The remainder of this paper is structured as follows. In Sec. \ref{sec:background}, we formulate the problem we aim to solve, as well as describe the use of GPs for modeling. In Sec. \ref{sec:MCPILCODF}, we detail the proposed algorithm, VF-MC-PILCO. Section \ref{sec:cartpoleValidation} illustrates the validation conducted on the simulated cart-pole benchmark. Section \ref{sec:UR5} reports the experiment performed on a simulated UR5 robot to test VF-MC-PILCO capacity of handling systems up to 6 DoF. Section \ref{sec:realExperiments} shows the results of the experiments with the two real mechanical systems. Finally, we draw conclusions in Section~\ref{sec:conclusions}.

\section{Background}\label{sec:background}
In this section, we first introduce the problem of MBRL on real mechanical systems. Then, we briefly discuss how Gaussian Process Regression (GPR) is usually used for modeling purposes.

\subsection{Problem Formulation}\label{sec:setting}
Consider a mechanical system with $d_{\boldsymbol{q}}$ degrees of freedom, and denote by $\boldsymbol{x}_{t}$ its state at time $t$. Typically, $\boldsymbol{x}_{t}$ is  defined as $\boldsymbol{x}_{t} = [\boldsymbol{q}_{t}^T,\boldsymbol{\dot{q}}_{t}^T]^T$, where $\boldsymbol{q}_{t} \in \mathbb{R}^{d_{\boldsymbol{q}}}$ and $\boldsymbol{\dot{q}}_{t}\in \mathbb{R}^{d_{\boldsymbol{q}}}$ are, respectively, the vector of joint positions and velocities. Assume that only $\boldsymbol{q}_t$ can be directly measured, whereas $\boldsymbol{\dot{q}}_t$  is not directly measurable, but instead must be estimated. We argue that this is a common scenario as mechanical systems are often equipped with position sensors such as encoders, but lack velocity sensors. More accurate velocity estimates can be obtained from the history of position measurements, exploiting both past as well as future samples. These kinds of estimation techniques are intrinsically acausal, hence they can be performed offline and used only for modeling. Thus, controllers must rely on causal online estimates, which are usually less accurate and affected by delays.

% The discrete-time dynamics of the system are represented by $\boldsymbol{x}_{t+1} = f(\boldsymbol{x}_{t}, \boldsymbol{u}_{t})$,  where $f(\cdot)$ is a (possibly stochastic) unknown transition function, and $\boldsymbol{u}_{t} \in \mathbb{R}^{d_{\boldsymbol{u}}}$ represents the control action.
The discrete-time dynamics of the system is given by $\boldsymbol{x}_{t+1} = f(\boldsymbol{x}_{t}, \boldsymbol{u}_{t}) + \boldsymbol{w}_{t}$,  where $f(\cdot)$ is an unknown transition function, $\boldsymbol{u}_{t} \in \mathbb{R}^{d_{\boldsymbol{u}}}$ represents the control action, and $\boldsymbol{w}_{t}\sim\mathcal{N}(0, \Sigma_{\boldsymbol{w}})$ models uncertainty. RL algorithms aim to learn how to accomplish a given task based on interaction data. The task is encoded by a cost function $c(\boldsymbol{x}_{t})$, defined to characterize the immediate penalty associated with being in state $\boldsymbol{x}_{t}$. Control actions are chosen from a policy $\boldsymbol{u}=\pi_{\boldsymbol{\theta}}(\boldsymbol{x})$, parameterized by $\boldsymbol{\theta}$. Then, the objective is to find the policy that minimizes the expected cumulative cost over $T$ time steps, starting from the initial state distribution $p(\boldsymbol{x}_{0})$, i.e.,
\begin{equation}\label{eq:expected_cost}
    J(\boldsymbol{\theta}) = \sum_{t=0}^T \mathbb{E}_{\boldsymbol{x}_{t}}\left[c\left(\boldsymbol{x}_{t}\right)\right]\text{.}
\end{equation}
An MBRL algorithm consists, generally, of the succession of several trials, i.e. attempts to solve the desired task, and each trial is structured in three main phases:

\begin{itemize}
    \item \textit{Model Learning}: data collected during previous interactions are used to learn a model of the system dynamics. At the beginning, first data are collected by applying an exploratory policy, e.g. random control actions;
    \item \textit{Policy Update}: the control policy is optimized in order to minimize an estimate of the cost $J(\boldsymbol{\theta})$ obtained by exploiting the trained model;
    \item \textit{Policy Execution}: the updated control policy is applied to the system and interaction data are stored.
\end{itemize}

In order to comply with the common conditions in real mechanical systems described above, we propose an MBRL algorithm to control mechanical systems without assuming to have neither measurements nor estimations of velocities.

\subsection{GPR for Model Learning}\label{sec:onestep_prediction}
Given a data set of state-action pairs measured during the interactions with the system, it is possible to use GPR to train a probabilistic model that approximates the unknown transition function $f(\cdot)$. A common strategy in the literature \cite{deisenroth2013gaussian,romeres2019semiparametrical} is to model the evolution of each state dimension with a distinct zero mean GP. Let us indicate with $x^{(i)}$ the \textit{i}-th component of the state, for $i \in \{1,\ldots, d_{\boldsymbol{x}}\}$ ($d_{\boldsymbol{x}}$ is the dimension of the state vector), and define $\tilde{\boldsymbol{x}}_{t} = [\boldsymbol{x}_{t}^T, \boldsymbol{u}_{t}^T]^T$. The \textit{i}-th GP takes $\tilde{\boldsymbol{x}}_{t}$ as input, and predicts $x^{(i)}_{t+1} - x^{(i)}_{t}$. The GPs are completely characterized by their kernel functions that represent our belief on the a priori covariance. A common choice is the Squared Exponential (SE) kernel, 
\begin{equation}\label{eq:SE}
    k_{SE}(\tilde{\boldsymbol{x}}_{t_j},\tilde{\boldsymbol{x}}_{t_k}) := \lambda^2 e^{-||\tilde{\boldsymbol{x}}_{t_j}-\tilde{\boldsymbol{x}}_{t_k}||^2_{\Lambda^{-1}}}\text{.}
\end{equation}
Given a data set of state-action pairs $\mathcal{D}$, the GPs provide a closed form expression of  $\boldsymbol{p}(\boldsymbol{x}_{t+1}|\tilde{\boldsymbol{x}}_t,\mathcal{D})$, the posterior distribution of the estimated state at time $t+1$. For further details about GPR and its application to dynamical system modeling, the readers can refer to \cite{williams2006gaussian}.

\section{Velocity-Free MC-PILCO}\label{sec:MCPILCODF}
Here we present the algorithm VF-MC-PILCO (\emph{Velocity Free Monte Carlo Probabilistic Inference for Learning COntrol}), whose objective is to solve the problem defined in Sec. \ref{sec:setting} without the need of performing any kind of velocity estimation. In fact, tuning effective estimators may be a tedious and complex operation, especially in the presence of high measurement noise. This might significantly compromise the MBRL algorithm, if not duly considered. VF-MC-PILCO circumvents these issues by adopting a VF formulation. Inspired by \cite{DFikin,DFRL}, we consider a VF model of the system dynamics, given in the following general form 
\begin{equation}\label{eq:generalDFmodel}
    \boldsymbol{q}_{t+1} = f_{\text{df}}(\boldsymbol{q}_{t}, \boldsymbol{q}_{t-1},\dots,\boldsymbol{q}_{t-m_{\boldsymbol{q}}},\boldsymbol{u}_{t},\dots,\boldsymbol{u}_{t-m_{\boldsymbol{u}}})\text{.} 
\end{equation}
The joint positions at the next time step are predicted based on the history of the past positions, from $t$ up to $t-m_{\boldsymbol{q}}$, and the history of applied control actions, from $t$ up to $t-m_{\boldsymbol{u}}$. Let $m_{\boldsymbol{q}}$ and $m_{\boldsymbol{u}}$ be called, respectively, the position memory and the control memory of the VF model. In this new VF framework, it is convenient to redefine the state of the system as $\boldsymbol{x}_{t} = [\boldsymbol{q}_{t}^T,\dots,\boldsymbol{q}_{t-m_{\boldsymbol{q}}}^T, \boldsymbol{u}_{t-1}^T,\dots,\boldsymbol{u}_{t-m_{\boldsymbol{u}}}^T]^T$.

In the following, we present the model learning and policy update phases of the VF-MC-PILCO algorithm, detailing how they have been adapted to the new VF formulation. 

\subsection{VF Model Learning} \label{sec:DF_model}
We employ the GPR framework of Sec. \ref{sec:onestep_prediction}, but instead of considering a full state representation with velocities, we train a VF GP model of form  \eqref{eq:generalDFmodel}. Let us denote with $q_t^{(i)}$ the position of the \textit{i}-th joint at time $t$, and define $\Delta^{(i)}_{q_t} = q^{(i)}_{t+1} - q^{(i)}_{t}$, for  $i \in \{1,\ldots, d_{\boldsymbol{q}}\}$. The evolution of $\Delta^{(i)}_{q_t}$ for all $i$ is modeled using a distinct GP, whose input depends upon $[\boldsymbol{q}_{t}^T,\dots,\boldsymbol{q}_{t-m_{\boldsymbol{q}}}^T, \boldsymbol{u}_{t}^T,\boldsymbol{u}_{t-1}^T,\dots,\boldsymbol{u}_{t-m_{\boldsymbol{u}}}^T]^T$. Trivially, the transition functions of  $\boldsymbol{q}_{t-1},\dots,\boldsymbol{q}_{t-m_{\boldsymbol{q}}}, \boldsymbol{u}_{t-1},\dots,\boldsymbol{u}_{t-m_{\boldsymbol{u}}}$ are deterministic and known.

Experimentally, we found it beneficial in terms of data efficiency (details in Sec. \ref{sec:inputVectAnalysis}) to rearrange GP input as 
\begin{equation}\label{eq:DF_GP_input}
    \tilde{\boldsymbol{x}}_{t} = \left[\boldsymbol{q}_t^T, \Delta_{\boldsymbol{q}_{t-1}}^T, \dots, \Delta_{\boldsymbol{q}_{t-m_{\boldsymbol{q}}}}^T, \boldsymbol{u}_{t}^T,\dots,\boldsymbol{u}_{t-m_{\boldsymbol{u}}}^T\right]^T\text{,}
\end{equation}
where $\Delta_{\boldsymbol{q}_{t-i}}=(\boldsymbol{q}_{t-i+1}-\boldsymbol{q}_{t-i})$, for $i=1,\dots,m_{\boldsymbol{q}}$, following the same notation used before when defining the GP targets.

In this way, we are providing the model with additional information about the rates of change observed inside the considered past position memory interval $m_{\boldsymbol{q}}$. Depending on the considered application, it may be convenient to further modify the GP input vector w.r.t. \eqref{eq:DF_GP_input}, in order to exploit particular characteristics of the considered quantities. For instance, we applied a \textit{sin-cos} expansion to angular quantities during some of the experiments presented in the next sections.

%As highlighted in \cite{DFikin,DFRL}, defining the input of the DF GPs as all the past  positions plus the past inputs might limit data efficiency, because of the large size of the input vector. To avoid that, it is convenient defining the input of the GPs by aggregating information on past positions. Indeed in \cite{DFikin}, they experimented with linear transformations of $\boldsymbol{q}_t \dots \boldsymbol{q}_{t-m_{\boldsymbol{q}}}$, with the coefficients treated as kernel hyperparameters. Due to the limited values of $m_{\boldsymbol{q}}$ needed in this work ($m_{\boldsymbol{q}}=1,2$), we adopt a simpler strategy. Information on past positions is condensed into the mean of the most recent consecutive differences. Then, the GP input becomes
%\begin{equation}\label{eq:diffGPinput}
%    \tilde{\boldsymbol{x}}_{t} = \left[\boldsymbol{q}_t^T, %\delta_{\boldsymbol{q}}(\boldsymbol{x}_{t},m_{\boldsymbol{q}})^T, \boldsymbol{u}_{t}^T,\dots,\boldsymbol{u}_{t-m_{\boldsymbol{u}}}^T\right]^T\text{,}
%\end{equation}
%with 
%\begin{equation}\label{eq:diffMean}
%    \delta_{\boldsymbol{q}}(\boldsymbol{x}_{t},m_{\boldsymbol{q}})=\frac{1}{m_{\boldsymbol{q}}} \sum_{j=1}^{m_{\boldsymbol{q}}} \left(\boldsymbol{q}_{t-j+1} - \boldsymbol{q}_{t-j}\right)\text{.}
%\end{equation}
%Experimentally, we found that providing this kind of information to the model is beneficial in terms of data efficiency and generalization.

\subsection{VF Particle-Based Policy Gradient}\label{sec:dfpolicyGradient}
The GP-based VF predictive model of Sec. \ref{sec:DF_model} is now employed to optimize the policy parameters $\boldsymbol{\theta}$ following a particle-based policy gradient strategy. VF-MC-PILCO computes $\hat{J}(\boldsymbol{\theta})$, an approximation of $J(\boldsymbol{\theta})$ in \eqref{eq:expected_cost} exploiting the posterior distribution $\boldsymbol{p}(\boldsymbol{x}_{t+1}|\tilde{\boldsymbol{x}}_t,\mathcal{D})$ defined by the GPs. Finally, it updates the $\boldsymbol{\theta}$ with a gradient-based procedure. 

\begin{figure}[t]
         \centering
         \includegraphics[width=\linewidth]{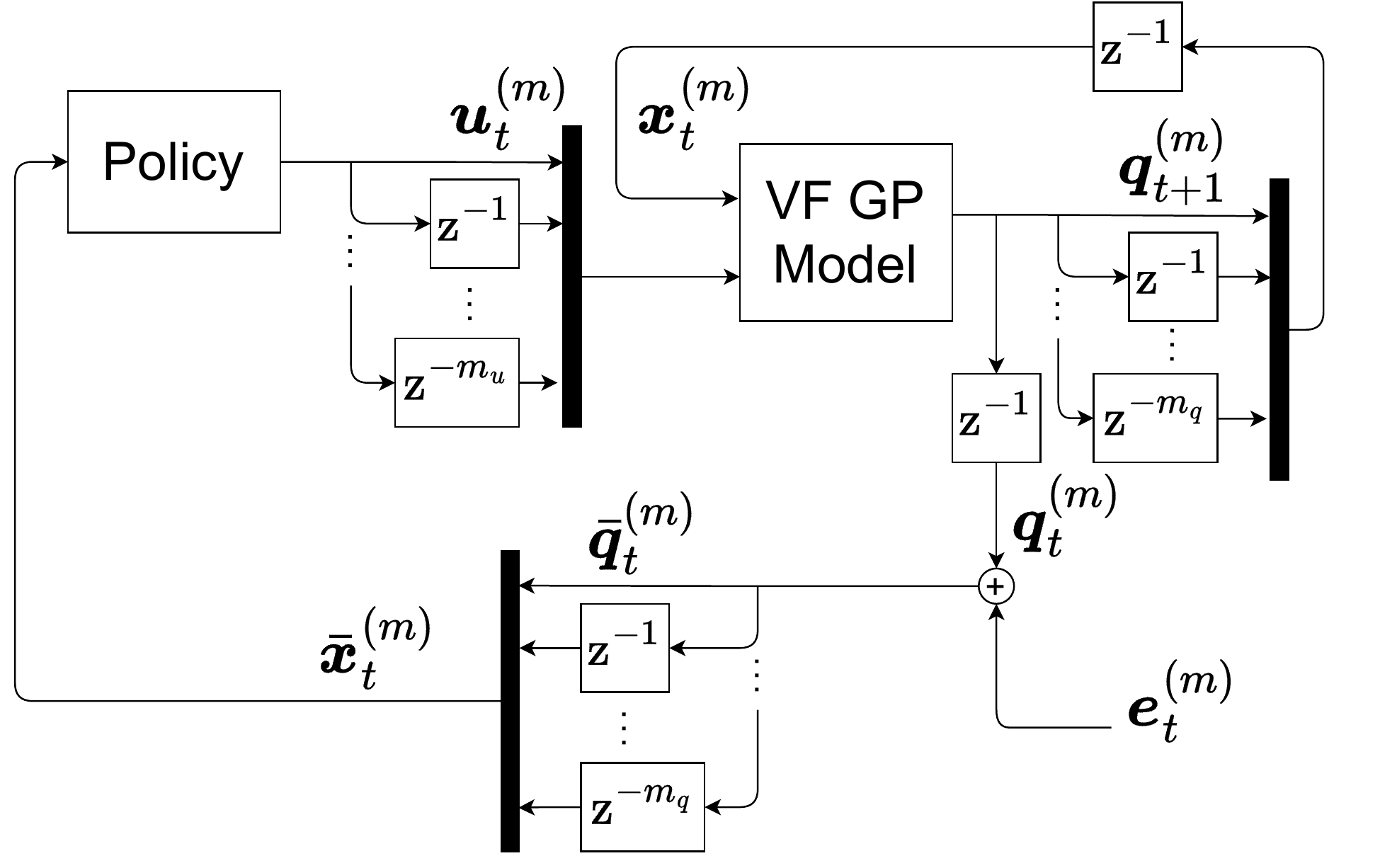}
         \caption{\small VF-MC-PILCO particles generation block schemes.}\label{fig:blockDF}
  \vspace{-3mm}
\end{figure}

The computation of $\hat{J}(\boldsymbol{\theta})$ entails the simulation of the effects of the policy $\boldsymbol{\pi_{\theta}}$ on $M$ independent state particles by cascading the one-step-ahead stochastic predictions. In particular, let $\boldsymbol{q}_t^{(m)}$, for $m=1,\dots,M$, represent the position of the $M$ state particles simulated by the VF GP model. Starting positions are sampled from a given distribution $p(\boldsymbol{q}_0)$. We assume that the system is not moving at $t=0$, i.e.,  $\boldsymbol{q}_0^{(m)}=\boldsymbol{q}_{-1}^{(m)}=\dots=\boldsymbol{q}_{-m_{\boldsymbol{q}}}^{(m)}$. At each time step $t$, in order to simulate the presence of measurement error, we corrupt the particle positions $\boldsymbol{q}_t^{(m)}$ with a fictitious noise $\boldsymbol{e}_t^{(m)}$, e.g. a zero mean Gaussian i.i.d. noise, obtaining a set of simulated measurements $\boldsymbol{\bar{q}}^{(m)}_t = \boldsymbol{q}^{(m)}_t + \boldsymbol{e}^{(m)}_t$. Then, for each particle, the policy $\boldsymbol{\pi_{\theta}}$  selects the next control actions $\boldsymbol{u}^{(m)}_t$ according to the history of the simulated measurements, $\boldsymbol{\bar{x}}_{t}^{(m)} = [\boldsymbol{\bar{q}}_{t}^{(m)T},\dots,\boldsymbol{\bar{q}}_{t-m_{\boldsymbol{q}}}^{(m)T}]^T$. Finally, the $M$ positions at the next time step, $t+1$, are simulated by forward sampling from the distributions derived by the VF GP model $\boldsymbol{p}(\boldsymbol{x}^{(m)}_{t+1}|\tilde{\boldsymbol{x}}^{(m)}_t,\mathcal{D})$ (for $m=1 \dots M$) with $\tilde{\boldsymbol{x}}^{(m)}_t$ defined for each particle $m$ as in \eqref{eq:DF_GP_input}. This procedure is iterated for \textit{T} time steps, obtaining \textit{M} different particle trajectories $\{\{\boldsymbol{x}_t^{(m)}\}_{m=1}^M\}_{t=0}^T$, that simulate the results of the policy. The particle generation procedure is depicted in the block scheme of Fig. \ref{fig:blockDF}. The sample mean of the costs incurred by the different particles provides an estimate of the expected cumulative cost, namely
\begin{equation}\label{eq:est_J}
    \hat{J}(\boldsymbol{\theta}) = \sum_{t=0}^T \left( \frac{1}{M}\sum_{m=1}^M c\left(\boldsymbol{x}_t^{(m)}\right)\right)\text{.}
\end{equation}
The computational graph resulting from \eqref{eq:est_J} allows us to compute $\nabla_{\boldsymbol{\theta}}\hat{J}(\boldsymbol{\theta})$, i.e., the gradient of $\hat{J}(\boldsymbol{\theta})$ w.r.t. $\boldsymbol{\theta}$, through backpropagation, exploiting the reparametrization trick \cite{kingma2013auto,rezende} to propagate the gradient through the stochastic operations. Finally, a stochastic gradient descent algorithm, e.g. Adam \cite{kingma2014adam}, can exploit the estimated gradient to update $\boldsymbol{\theta}$.

\subsection{Policy structure}\label{subsec:policy_structure}
We considered an RBF network policy with outputs limited by a hyperbolic tangent function, properly scaled. We call this function \textit{squashed-RBF-network}, and it is expressed as
\begin{equation}\label{eq:policy}
    \pi_{\boldsymbol{\theta}}(\boldsymbol{x}^*) = u_{max}\;\text{tanh} \left(\frac{1}{u_{max}}\sum_{i=1}^{n_b} w_i e^{||\boldsymbol{a}_i-\boldsymbol{x}^*||_{\Sigma_{\pi}}^2}\right)\text{,}
\end{equation}
The input vector of the policy is defined as
\begin{equation}\label{eq:diffPolicyInput}
\boldsymbol{x}^*_t=\left[\boldsymbol{q}_t^T, \Delta_{\boldsymbol{q}_{t-1}}^T, \dots, \Delta_{\boldsymbol{q}_{t-m_{\boldsymbol{q}}}}^T\right]^T,
\end{equation}
where we are providing the policy with the same consecutive differences of position measures used for GP input in \eqref{eq:DF_GP_input}. 

The policy parameters are $\boldsymbol{\theta} = \left\{\boldsymbol{w}, A,\Sigma_{\pi}\right\}$, where $\boldsymbol{w}=[w_1\dots w_{n_b}]$ and $A=\left\{\boldsymbol{a}_1\dots\boldsymbol{a}_{n_b}\right\}$ are, respectively, the weights and the centers of the $n_b$ basis functions, while ${\Sigma_{\pi}}$ is a diagonal matrix that determines theirs shapes. The maximum control $u_{max}$ is constant and depends on the application.  It is worth mentioning that VF-MC-PILCO is not restricted to this particular choice of policy function.

\section{Simulated Experiment: Cart-pole Swing-Up}\label{sec:cartpoleValidation}
As a preliminary validation, we tested VF-MC-PILCO on a simulated cart-pole swing-up task to analyze its performance under different setups. We compare the proposed approach with the s.o.t.a. MBRL algorithm specifically designed to deal with partial state measurability of real mechanical systems, MC-PILCO4PMS \cite{amadio2021tro}. MC-PILCO4PMS follows a particle-based policy gradient framework similar to the one depicted in Sec. \ref{sec:dfpolicyGradient}, but, differently from the proposed VF-MC-PILCO, it works with velocity estimates by simulating not only the evolution of the system state but also the evolution of the estimated state, which entails modeling the measurement system and the implemented online filters. Notice that the implementation of MC-PILCO4PMS could be in some cases complex or time-consuming, due to its requirement to reproduce the online filtering procedure inside the policy optimization phase, and the need to adopt a different offline filter for model learning. This is the limitation that the proposed method aims to solve. Both algorithms have been implemented in Python\footnote{\url{https://www.merl.com/research/license/MC-PILCO}}, exploiting the PyTorch library \cite{paszke2019pytorch}.

Now, let us briefly describe the characteristics of the simulated scenario. Let $p_t$ and $\alpha_t$ be, respectively, the position of the cart and the angle of the pole at time step $t$, hence $\boldsymbol{q}_t=[p_t, \alpha_t]^T$. The target configurations corresponding to the pendulum swing-up are given by $p^{des}=0$ [m] and $\vert \alpha^{des} \vert = \pi$ [rad]. The cart-pole starts from $\theta_0 = 0$ [rad] and $p_0=0$ [m]. The control action is the force applied to the cart, and the system is controlled at 30 [Hz]. We considered a Gaussian measurement noise with standard deviation of $10^{-3}$ [m] for positions and $2\cdot 10^{-3}$ [rad] for angles.

The GPs of the VF model are equipped with the SE kernel described in \eqref{eq:SE}. The policy adopted is a \textit{squashed-RBF-network} policy with $n_b=200$ basis functions and $u_{max}=10$ [N]. The number of particles is set to $M=400$ during policy optimization. In order to avoid singularities due to the angles, we replaced, in both the model inputs $\tilde{\boldsymbol{x}}_{t}$ defined in \eqref{eq:DF_GP_input} and policy input $\boldsymbol{x}^*$ defined in \eqref{eq:diffPolicyInput}, occurrences of $\alpha_t$ with $sin(\alpha_t)$ and $cos(\alpha_t)$. Exploration data were collected by random actions, obtained by filtering Gaussian white noise with cut-off frequency $1.5$ [Hz]. The cost function is 
\begin{equation}\label{eq:abs_cost_cartpole}
   c(\boldsymbol{x}_t) = 1-\text{exp}\left(-\left(\frac{|\alpha_t|-\pi}{l_{\alpha}}\right)^2 -\left(\frac{p_t}{l_p}\right)^2\right),
\end{equation}
where $l_{\alpha}=3$ and $l_p=1$ define the shape of $c(\cdot)$. The absolute value on $\alpha_t$ is needed to allow different swing-up solutions to both the equivalent target pole angles $\pi$ and $-\pi$.

The objective is to analyze different VF-MC-PILCO configurations and compare their performance with the results obtained by MC-PILCO4PMS, as a benchmark. We analyzed the results obtained in 50 distinct experiments, consisting of 5 trials of length 3 seconds, varying the random seed each time. In this way, it is possible to evaluate the robustness of the algorithm to different exploration trajectories and policy initialization, as well as to different noise realizations. In particular, we investigate the effects that different position and control memories, $m_{\boldsymbol{q}}$ and $m_{\boldsymbol{u}}$, have on modeling and policy learning. We studied four different VF-MC-PILCO configurations, choosing the value of $m_{\boldsymbol{q}}$ between 1 and 2, and $m_{\boldsymbol{u}}$ between 0 and 1. In the following, we will refer to these different alternatives with the symbol $VF^{m_{\boldsymbol{u}}}_{m_{\boldsymbol{q}}}$.

\subsection{Modeling results}
\begin{figure}[t]
\centering
  \includegraphics[width=\linewidth]{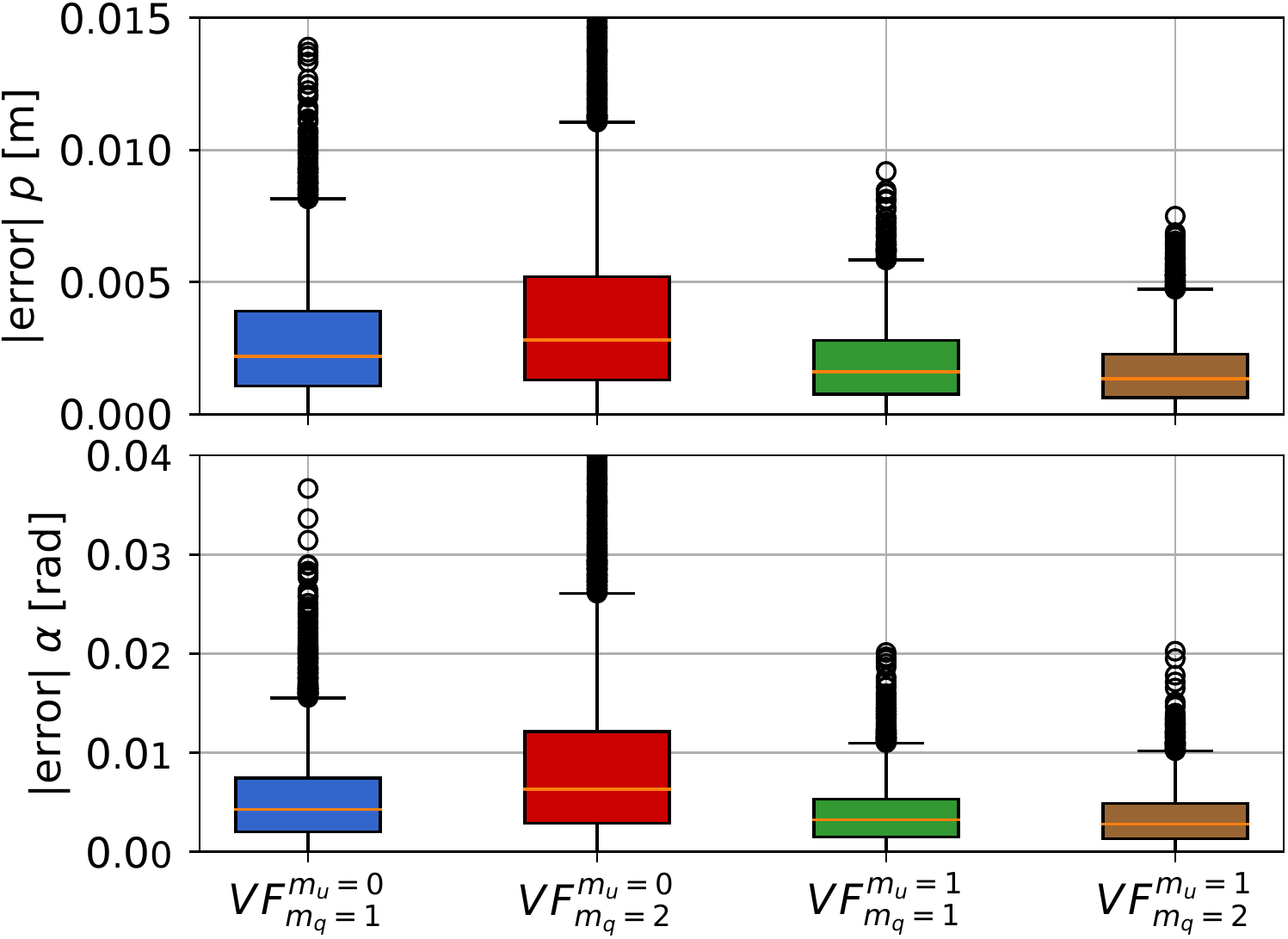}
  \caption{\small Absolute $p$ and $\alpha$ prediction errors obtained by different VF GP models at trial 5 in the simulated cart-pole experiments.}
  \label{fig:boxPlot}
  \vspace{-3mm}
\end{figure}

We compared the accuracy of the different VF GP models by looking at the absolute values of the prediction errors observed on the data registered at the last trial in all 50 experiments. Models were trained using  all the data collected up to that trial. Fig. \ref{fig:boxPlot} reports the results by means of box plots, showing median values, confidence intervals, and outliers. The results show that  the presence of input history, $\boldsymbol{u}_{t-1}$, as part of the GP inputs is beneficial, and within this choice, see models with $m_{\boldsymbol{u}}=1$, the best results are obtained by $VF^{m_{\boldsymbol{u}}=1}_{m_{\boldsymbol{q}}=2}$. On the other hand, the greater errors and the significant number of outliers obtained by models with $m_{\boldsymbol{u}}=0$ seem to indicate that these kinds of setups are not fully capable of fitting the registered position changes. Also, it appears that using a longer position memory leads to an improvement of prediction accuracy only when the control memory is $m_{\boldsymbol{u}}=1$. We can conclude that it seems beneficial to provide VF GP models with information about past control actions ($m_{\boldsymbol{u}}=1$) for fitting the system dynamics without relying on velocity. 

\subsection{Policy learning results}
\begin{figure}[t]
\centering
  \includegraphics[width=\linewidth]{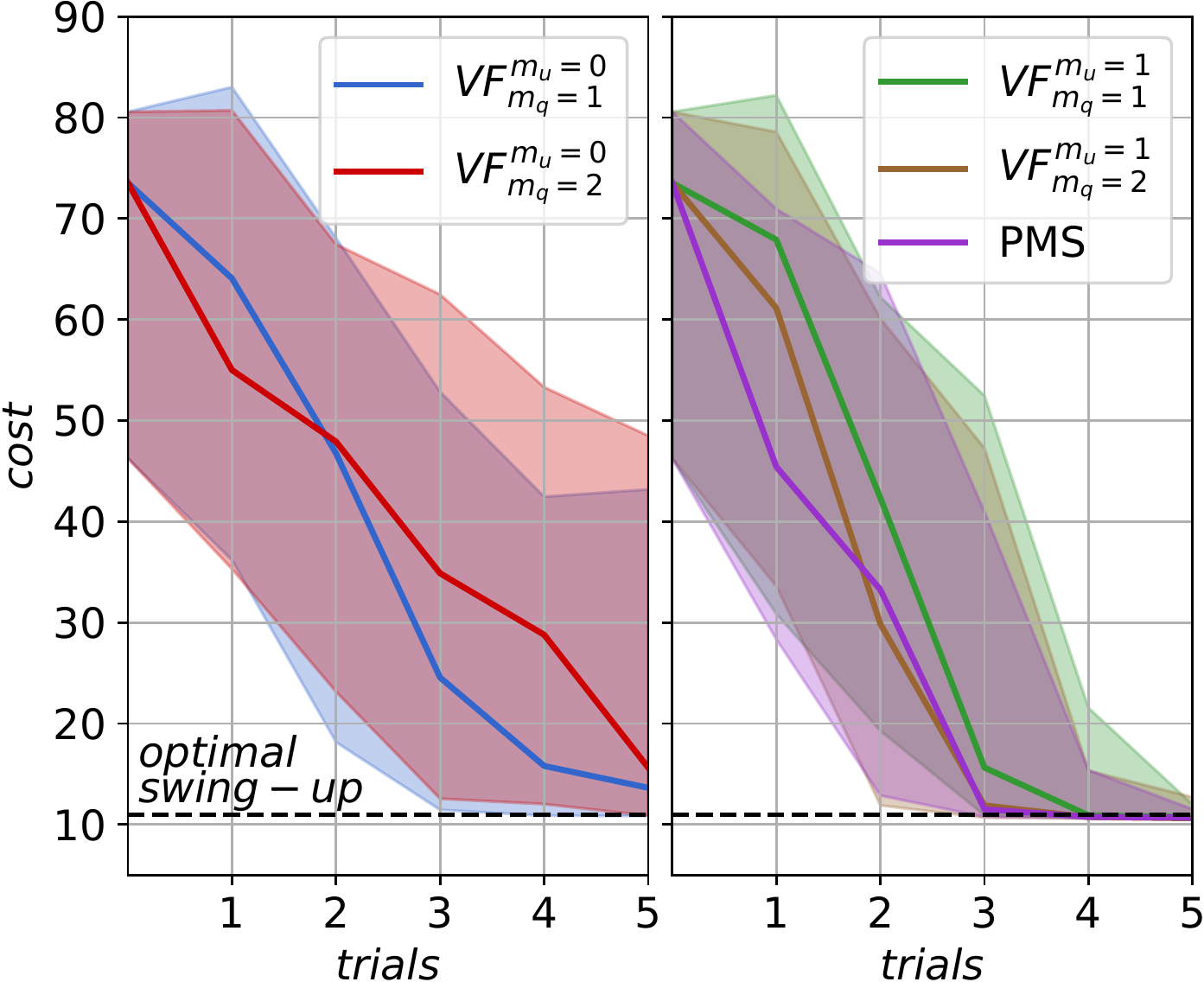}
  \caption{\small Cumulative costs registered during simulated cart-pole experiments from the four considered VF-MC-PILCO setups and the MC-PILCO4PMS benchmark (indicated by the shorthand PMS). The observed success rates are presented in the table below.}
  \smallskip
  \begin{tabular}{|l|l|l|l|l|l|}
  \hline
  & \small{Trial 1} & \small{Trial 2} & \small{Trial 3} & \small{Trial 4} & \small{Trial 5} \\ \hline
  \small{$VF^{m_{\boldsymbol{u}}=0}_{m_{\boldsymbol{q}}=1}$} & \small{0\%} & \small{14\%} & \small{34\%} & \small{46\%} & \small{56\%} \\ \hline
  \small{$VF^{m_{\boldsymbol{u}}=0}_{m_{\boldsymbol{q}}=2}$} & \small{0\%} & \small{10\%} & \small{18\%} & \small{28\%} & \small{52\%} \\ \hline
  \small{$VF^{m_{\boldsymbol{u}}=1}_{m_{\boldsymbol{q}}=1}$} & \small{0\%} & \small{8\%} & \small{52\%} & \small{86\%} & \small{96\%} \\ \hline
  \small{$VF^{m_{\boldsymbol{u}}=1}_{m_{\boldsymbol{q}}=2}$} & \small{0\%} & \small{20\%} & \small{73\%} & \small{93\%} & \small{100\%} \\ \hline
  \small{PMS} & \small{0\%} & \small{14\%} & \small{82\%} & \small{98\%} & \small{96\%} \\ \hline
 \end{tabular}
  \label{fig:cartpoleCost}
  \vspace{-3mm}
\end{figure}
In this section, we evaluate the performance of the control policies learned by the different VF-MC-PILCO setups and by MC-PILCO4PMS. Notice that MC-PILCO4PMS achieved results comparable to or better than other state-of-the-art GP-based MBRL algorithms, see \cite{amadio2021tro}. The cumulative costs and success rates obtained at each trial in the 50 experiments are reported in Fig. \ref{fig:cartpoleCost}. In the two plots, the cumulative cost is reported in terms of median values and confidence intervals defined by the 5-th and 95-th percentiles. As one would expect, the worse modeling results of $VF^{m_{\boldsymbol{u}}=0}_{m_{\boldsymbol{q}}=1}$ and $VF^{m_{\boldsymbol{u}}=0}_{m_{\boldsymbol{q}}=2}$ lead to an unsatisfactory policy learning. These VF-MC-PILCO setups manage to complete a successful swing-up only in, approximately, half of the cases. On the other hand, when using $m_{\boldsymbol{u}}=1$, VF-MC-PILCO is able to robustly find an optimal solution for the task by trial 5. In particular, the performance of $VF^{m_{\boldsymbol{u}}=1}_{m_{\boldsymbol{q}}=2}$ are almost equivalent to the results of MC-PILCO4PMS. This result confirms the effectiveness of the proposed method: with less information, as we are not manually tuning any velocity estimator, VF-MC-PILCO achieves state-of-the-art performance. For the user, this corresponds to less effort and a more general method without compromising significantly the performance.

\subsection{Analysis of input vector structure}\label{sec:inputVectAnalysis}
\begin{figure}[t]
\centering
  \includegraphics[width=\linewidth]{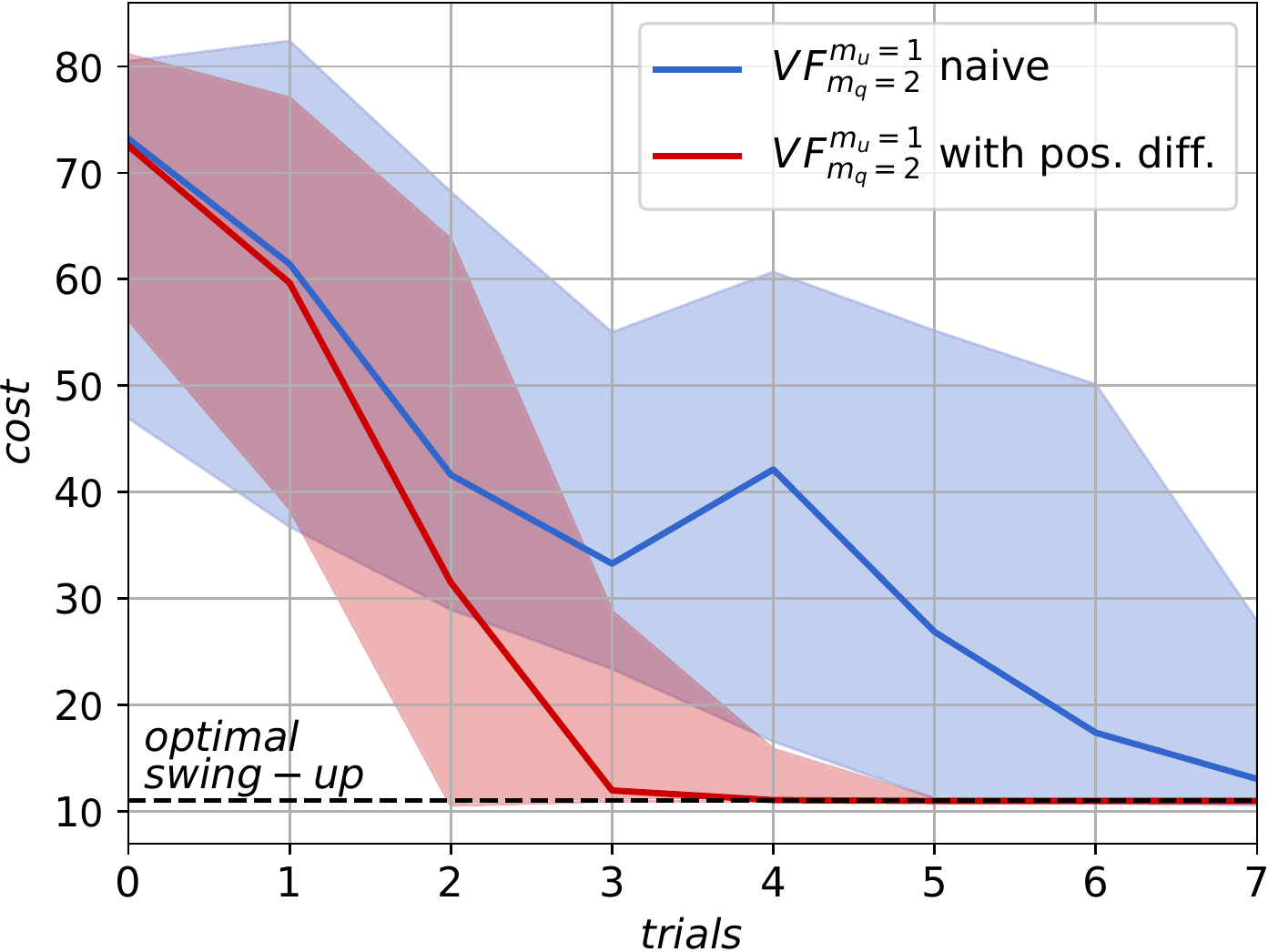}
  \caption{\small Cumulative costs registered by VF-MC-PILCO with different GP and policy input structures that include past positions either as consecutive differences ($\Delta_{\boldsymbol{q}_{t-1}}, \dots, \Delta_{\boldsymbol{q}_{t-m_{\boldsymbol{q}}}}$) or directly ($\boldsymbol{q}_{t-1}, \dots, \boldsymbol{q}_{t-m_{\boldsymbol{q}}}$).}
  \label{fig:DiffHistory_VS_Naive}
  \vspace{-3mm}
\end{figure}

Before concluding this section, we would like to analyze the reasons behind the decision to use \eqref{eq:DF_GP_input} and \eqref{eq:diffPolicyInput} as GP and policy input vectors, respectively. In this respect, we compared the results obtained by VF-MC-PILCO with position memory $m_{\boldsymbol{q}}=2$ and control memory $m_{\boldsymbol{u}}=1$ using two different structures for the input vectors. The first employs directly the history of positions and actions up to time step $t$ as GP input, e.g., $\tilde{\boldsymbol{x}}_{t}=[\boldsymbol{q}_{t}^T,\dots,\boldsymbol{q}_{t-m_{\boldsymbol{q}}}^T, \boldsymbol{u}_{t}^T,\dots,\boldsymbol{u}_{t-m_{\boldsymbol{u}}}^T]^T$, and the history of positions as policy input, e.g., $\boldsymbol{x}^*_t=[\boldsymbol{q}_{t}^T,\dots,\boldsymbol{q}_{t-m_{\boldsymbol{q}}}^T]^T$. The second version is the one employed previously with GP input and policy input defined as in \eqref{eq:DF_GP_input} and \eqref{eq:diffPolicyInput}, respectively. To distinct the two implementations, we labeled the first as $VF^{m_{\boldsymbol{u}}=1}_{m_{\boldsymbol{q}}=2}$ \textit{naive}, and the second  $VF^{m_{\boldsymbol{u}}=1}_{m_{\boldsymbol{q}}=2}$ \textit{with position differences}. We analyzed the results obtained by these two setups in 50 distinct experiments, consisting of 7 trials of length 3 seconds, varying the random seed each time. The obtained cumulative costs are reported in Fig. \ref{fig:DiffHistory_VS_Naive} in terms of median values and 5-95 percentile ranges. 

It is clear that providing information about the rate of change of position measures, by using \eqref{eq:DF_GP_input} and \eqref{eq:diffPolicyInput} as input vectors, greatly improves the data efficiency of VF-MC-PILCO algorithm. In fact, the $VF^{m_{\boldsymbol{u}}=1}_{m_{\boldsymbol{q}}=2}$ \textit{naive} implementation (that uses directly the history of position and controls) shows a much slower convergence, reaching a 79\% success rate only at trial 7. On the other hand, $VF^{m_{\boldsymbol{u}}=1}_{m_{\boldsymbol{q}}=2}$ \textit{with position differences} is able to always find a solution by trial 5. 

This result underlines the importance of the information carried out by the differences between consecutive measured positions. Without this information, the model needs more data to correctly capture the dynamics of the system relying only on positions. Through input vectors \eqref{eq:DF_GP_input} and \eqref{eq:diffPolicyInput}, we are able to provide the model with knowledge about a sort of velocity, without requiring any kind of filtering procedure.

\section{Simulated Experiment: UR5 Robot Control }\label{sec:UR5}
The objective of this experiment is to test VF-MC-PILCO in a more complex system with a higher DoF. We used VF-MC-PILCO to learn a joint-space torque controller for a UR5, a robotic manipulator with 6 DoF, simulated in MuJoCo \cite{todorov2012mujoco}, assuming to measure only joint angles and not velocities. Measurements are perturbed by the presence of white Gaussian noise with a standard deviation of $10^{-3}$. Let us denote with $\boldsymbol{q}_t \in \mathbb{R}^6$ the joint angles and with $\boldsymbol{u}_t \in \mathbb{R}^6$ the applied torques. Our objective is to learn a VF control policy able to follow a desired trajectory $\{\boldsymbol{q}^r_t\}_{t=1}^T$. Let $\boldsymbol{e}_t = \boldsymbol{q}^r_t-\boldsymbol{q}_t$ denote the position error at time $t$. VF-MC-PILCO memories were set to $m_{\boldsymbol{q}}=2$ and $m_{\boldsymbol{u}}=1$, hence the VF state of the system at time step $t$ is defined as $\boldsymbol{x}_t=[\boldsymbol{q}_t^T, \boldsymbol{q}_{t-1}^T, \boldsymbol{q}_{t-2}^T, \boldsymbol{u}_{t-1}^T]^T$. The GP input vector was defined applying a \textit{sin-cos} expansion of angular quantities as $\tilde{\boldsymbol{x}}_t = [sin(\boldsymbol{q}_t)^T, cos(\boldsymbol{q}_t)^T, \Delta_{\boldsymbol{q}_{t-1}}^T, \Delta_{\boldsymbol{q}_{t-2}}^T, \boldsymbol{u}_t^T, \boldsymbol{u}_{t-1}^T]^T$.

\begin{figure}[h]
     \centering
     \includegraphics[width=\linewidth]{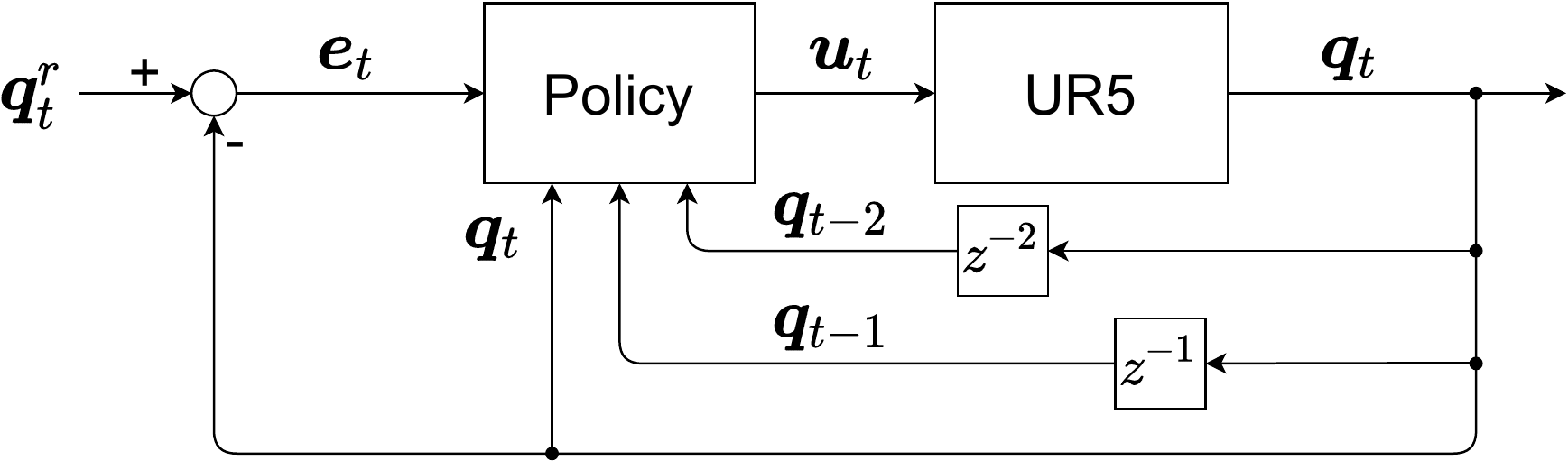}
     \caption{\small VF-MC-PILCO control scheme for the simulated UR5.}
     \label{fig:UR5_DF_ctrl_scheme}
  \vspace{-3mm}
\end{figure}

The policy adopted was a multi-output \textit{squashed-RBF-network} with $n_b=400$ basis functions and $u_{max}=1$ [N$\cdot$m] for all the joints. $M=200$ particles were used during optimization. The policy takes in input the vector $\boldsymbol{x}_t^* = [sin(\boldsymbol{q}_t)^T, cos(\boldsymbol{q}_t)^T, \Delta_{\boldsymbol{q}_{t-1}}^T, \Delta_{\boldsymbol{q}_{t-2}}^T, \boldsymbol{e}_t^T]^T$. 
Fig.~\ref{fig:UR5_DF_ctrl_scheme} represents the overall control scheme.

In this experiment, we considered a control horizon of 4 seconds with a sampling time of 0.02 seconds. The reference trajectory has been calculated to make the end-effector draw a circle in the X-Y operational space. The initial exploration used to initialize the VF GP model is provided by a poorly-tuned PD controller (for which we estimated velocity by backward differentiation). We used $M=200$ of particles for gradient estimation and considered the following cost,
\begin{equation*}
    c(\boldsymbol{x}_t)=1-\text{exp}\left( ||\boldsymbol{q}^r_t-\boldsymbol{q}_t||^2 \right).
\end{equation*}

\begin{figure}[t]
     \centering
     \includegraphics[width=0.70\linewidth]{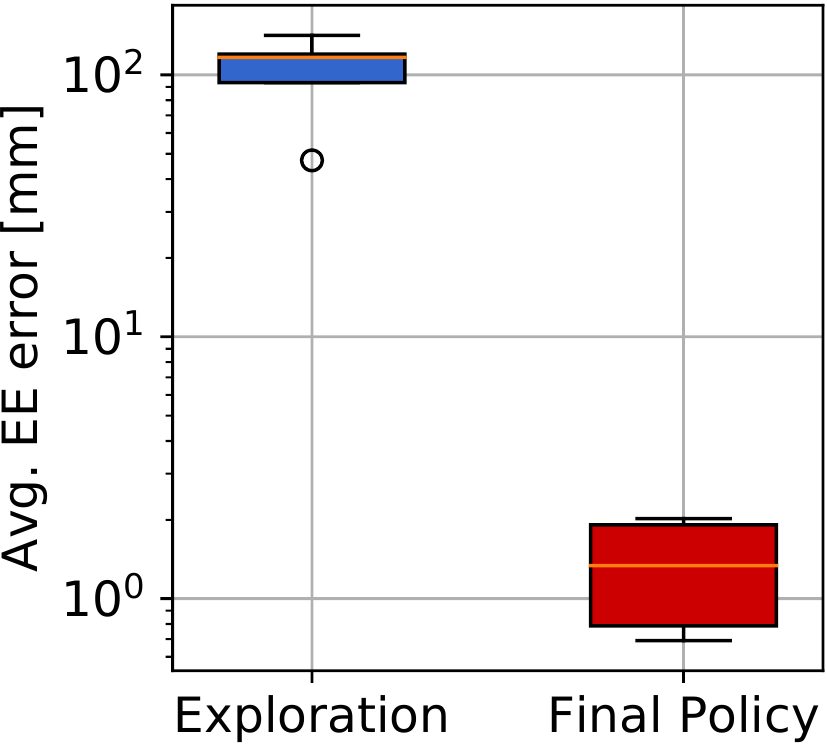}
     \caption{\small Average end-effector position tracking errors obtained during the exploratory phases and by the control policy learned at trial 3.}\label{fig:UR5_boxplot}
  \vspace{-3mm}
\end{figure}

\begin{figure}[t]
     \centering
     \includegraphics[width=0.95\linewidth]{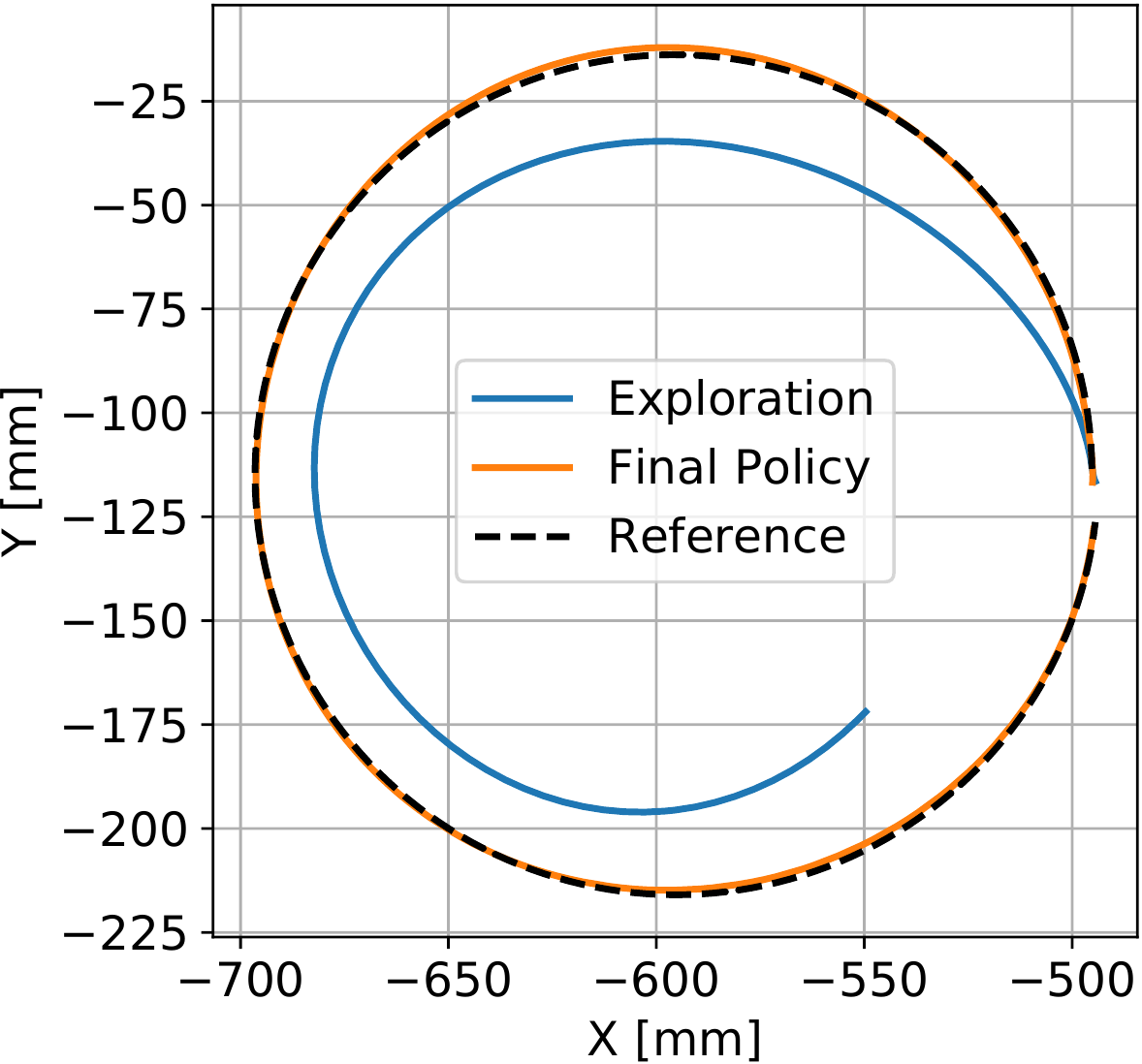}
     \caption{\small Example of explorative and final end-effector trajectories.}\label{fig:DF_UR5_trajectories}
  \vspace{-3mm}
\end{figure}

The experiment was repeated 10 different times, varying the random seed and the initial exploration trajectories, obtained each time  by using random PD gains, uniformly sampled from $K_P \sim \mathcal{U}(0.5,2)$ and $K_D \sim \mathcal{U}(0.01,0.2)$. VF-MC-PILCO managed to learn an effective control policy by the third trial in all the repetitions, with average positioning errors not superior to 2 [mm].
The average end-effector tracking errors obtained are reported in Fig. \ref{fig:UR5_boxplot}, where results are given by means of box plots. 
Fig. \ref{fig:DF_UR5_trajectories} shows an example of exploratory and final trajectories, taken from one of the conducted tests.

% VF-MC-PILCO managed to learn an effective control policy by the third trial in all the repetitions.
% \begin{wrapfigure}{r}{0.6\linewidth}
%  \centering
%  \includegraphics[width=\linewidth]{pictures/boxplot_UR5_DF.pdf}
%  \caption{\small Average end-effector position tracking errors obtained during the exploratory phases and by the VF-MC-PILCO control policy learned at trial 3.}\label{fig:UR5_boxplot}
% \end{wrapfigure}
% The average end-effector tracking errors obtained are reported in Fig. \ref{fig:UR5_boxplot}, where results are given by means of box plots. Indicatively, we observed average positioning errors not superior to 2 [mm]. Finally, in Fig. \ref{fig:DF_UR5_trajectories} we reported an example of exploratory and final trajectories, taken from one of the conducted test.

\section{Experiments on Real Mechanical Systems}\label{sec:realExperiments}
\begin{figure}[t]
\centering
  \includegraphics[width=\linewidth]{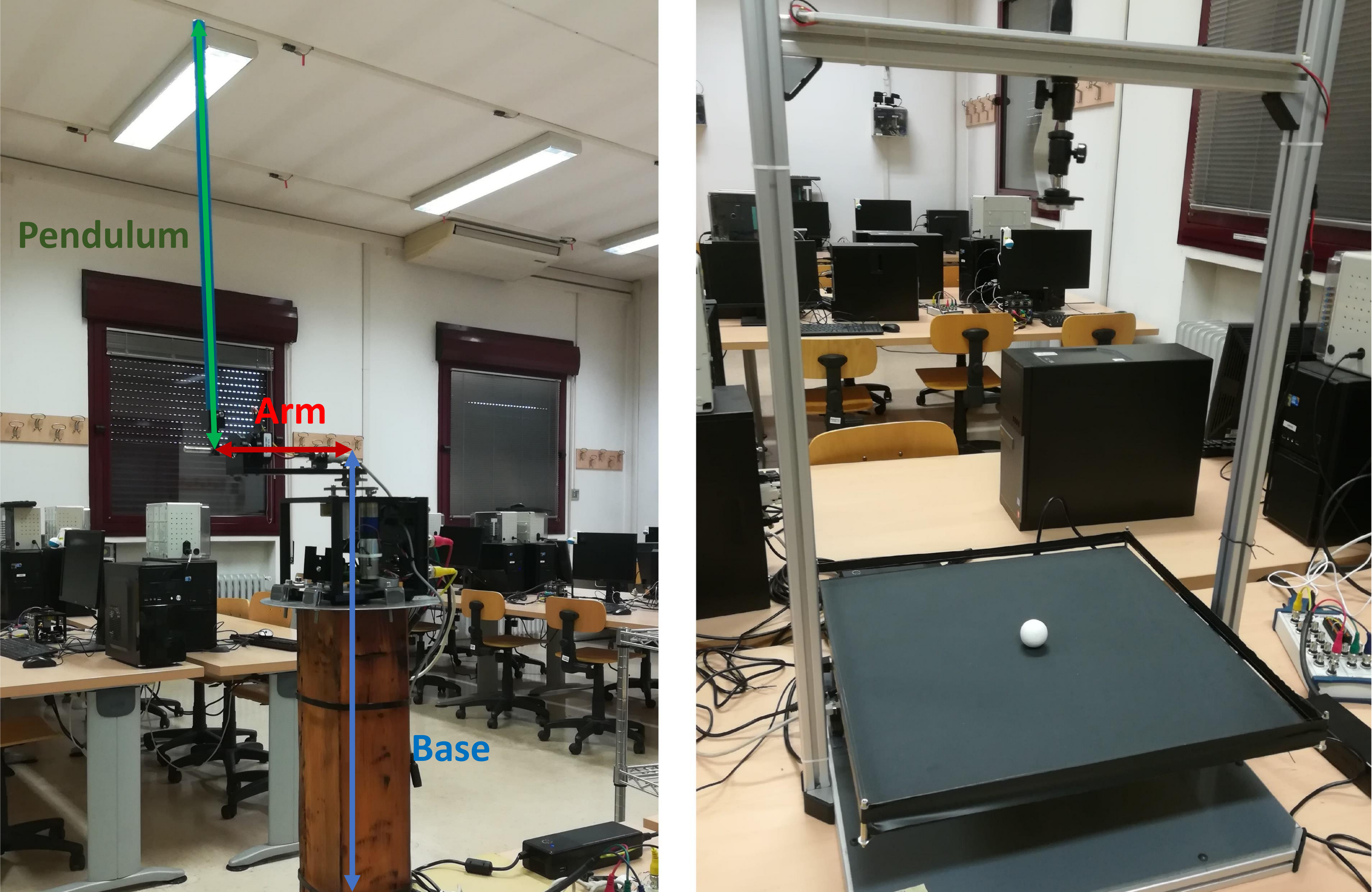}
  \caption{\small (Left) Furuta pendulum. (Right) Ball-and-plate system.}
  \label{fig:real_systems}
  \vspace{-3mm}
\end{figure}

In this section, we report the results obtained by VF-MC-PILCO when applied to real systems. In particular, we experimented on two benchmark systems: a Furuta pendulum, and a ball-and-plate (Figure \ref{fig:real_systems})\footnote{A video of the experiments on real mechanical systems is available at the following link \url{https://youtu.be/Hx3Y1Ib-6Tc}.}. The objective is to compare the performance obtained by VF-MC-PILCO in these two setups with the results of MC-PILCO4PMS reported in \cite{amadio2021tro}.
\subsection{Furuta pendulum}
The Furuta pendulum \cite{cazzolato2011furuta} is a popular nonlinear control benchmark composed of two revolute joints and three links (see Fig. \ref{fig:real_systems}, left). It is an under-actuated system as only the horizontal joint is actuated by a DC servomotor. The two angles are measured by optical encoders with  4096 [ppr]. The control action is the motor voltage, and its maximum allowed value is 10 [V]. Let the pose at time step $t$ be $\boldsymbol{q}_t = [\alpha^h_t, \alpha^v_t]^T$, where $\alpha^h_t$ is the angle of the horizontal joint and $\alpha^v_t$ the angle of the vertical joint attached to the pendulum. The objective is to learn how to swing-up the pendulum and stabilize it in the upward equilibrium ($\alpha_t^v=\pm\pi$ [rad]) with $\alpha_t^h=0$ [rad], starting from $\boldsymbol{q}_0 = [0, 0]^T$. The trial length is 3 [s] and the system is controlled at 30 [Hz]. The cost is defined as
\begin{equation}\label{eq:furuta cost}
   c(\boldsymbol{x}_t)=1-\text{exp}\left( -\left(\frac{\alpha_t^h}{2}\right)^2  -\left(\frac{|\alpha_t^v|-\pi}{2}\right)^2 \right) +c_{b}(\alpha_t^h),
\end{equation}
with
\begin{align*}
    c_{b}(\alpha_t^h) =& \frac{1}{1+\text{exp}\left(-10\left(-\frac{3}{4}\pi-\alpha^h_t \right)\right)}\\
    &+\frac{1}{1+\text{exp}\left(-10\left(\alpha^h_t-\frac{3}{4}\pi \right)\right)}\text{.}
\end{align*}
The first part of the function in \eqref{eq:furuta cost} aims at driving the two angles towards the target, while $c_{b}(\alpha_t^h)$ penalizes solutions that push the horizontal joint beyond a certain safety threshold.

In this scenario, we used position memory $m_{\boldsymbol{q}}=2$ and control memory $m_{\boldsymbol{u}}=1$. We equipped the VF GP model with an SE kernel and adopted a \textit{squashed-RBF-network} with $n_b=200$ basis functions as control policy. $M=400$ particles were simulated during policy optimization. We replaced, in both GP inputs $\tilde{\boldsymbol{x}}_{t}$ and policy input $\boldsymbol{x}^*$, occurrences of $\alpha^h_t$ and $\alpha^v_t$ with their \textit{sin-cos} expansion, as previously done in the simulated cart-pole case. The exploration trajectory has been obtained using as input a sum of ten cosine waves of random frequencies and the same amplitudes. The presence of quantization errors was simulated during particles generation by corrupting predicted angles with a uniform fictitious measurement noise $\mathcal{U}(\frac{-\pi}{4096},\frac{\pi}{4096})$ [rad]. 

\begin{figure}[t]
\centering
  \includegraphics[width=0.95\linewidth]{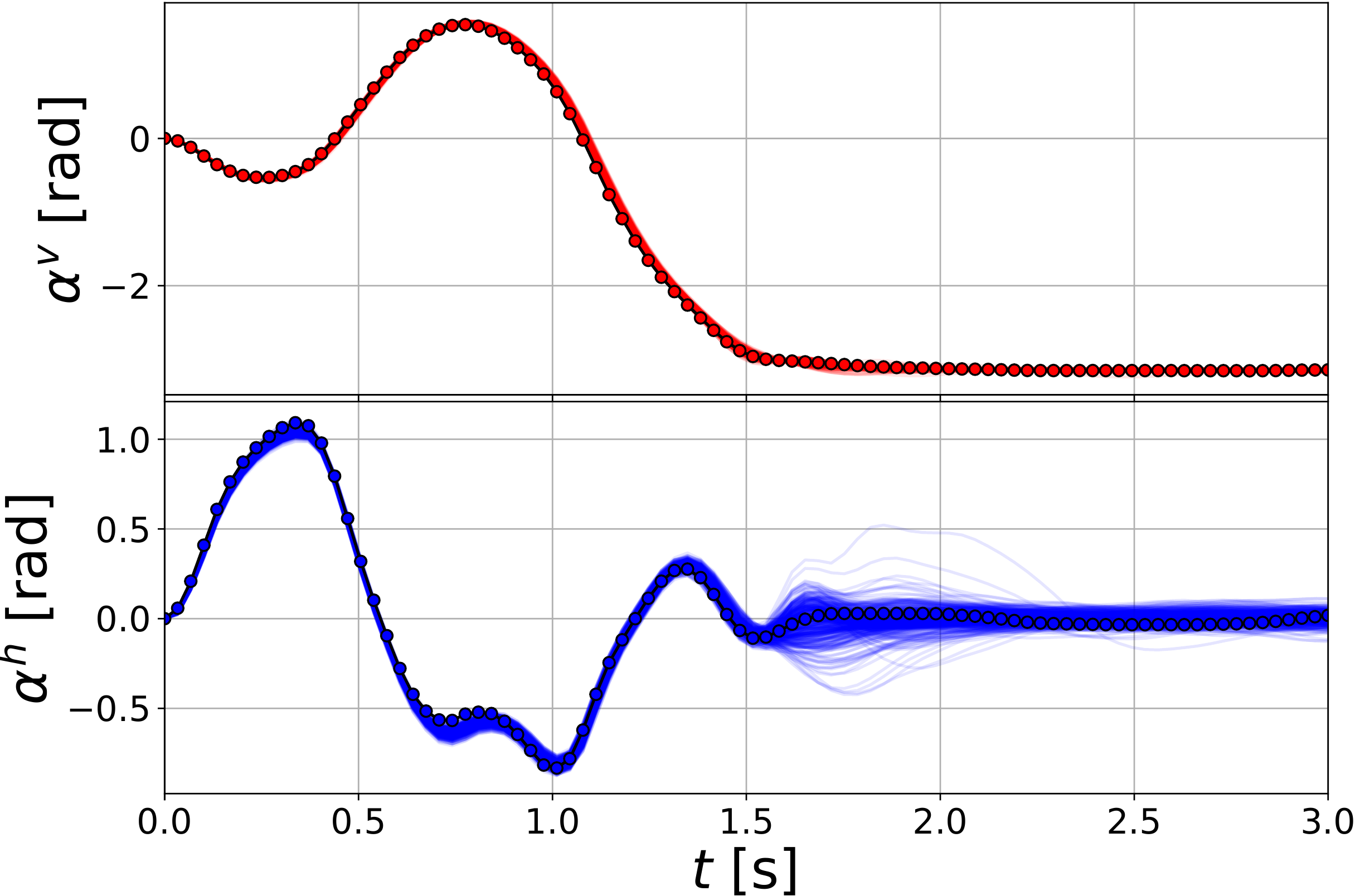}
  \caption{\small Real swing-up trajectory (bullets) and particles prediction (shaded lines) obtained by VF-MC-PILCO at trial 6 of the Furuta pendulum experiment.}
  \label{fig:fp_results}
  \vspace{-3mm}
\end{figure}

VF-MC-PILCO learned how to swing-up the Furuta pendulum at trial 6, i.e. after 18 seconds of experience. That is the same result obtained by MC-PILCO4PMS when using the SE kernel. Hence, the VF approach showed no particular differences in terms of data  efficiency when compared with an approach that makes use of velocity estimates.
In Fig. \ref{fig:fp_results}, we report the successful swing-up performed by VF-MC-PILCO at trial 6, together with the particles predicted by the VF GP model, simulating the effects of the same control policy. Notice how the particles' trajectories resemble almost perfectly the real behaviour of the two angles.

\subsection{Ball-and-plate}
The ball-and-plate system is composed of a square plate that can be tilted in two orthogonal directions, and a ball that is free to roll over it (see Fig. \ref{fig:real_systems}, right). A camera is placed on top of the system to track the ball and measure its position on the plate, with a precision of one millimeter. Let, at time $t$, $(b^x_t,b^y_t)$ be the position of the center of the ball, while $\alpha^{(1)}_t$ and $\alpha^{(2)}_t$ are the angles of the two motors tilting the plate. Thus, $\boldsymbol{q}_t = [b^x_t,b^y_t,\alpha^{(1)}_t,\alpha^{(2)}_t]^T$. The drivers of the motors allow only position control and do not provide feedback about the motors' angles. To keep track of them, we defined the control actions as the difference between two consecutive reference values sent to the motor, and we limited the maximum input to a sufficiently small value, i.e. 4 [deg], such that the motor controllers are able to reach the target within the sampling time. Then, as a first approximation, the reference angles, and the actual motor angles coincide, and we have $u_t^{(1)} = \alpha^{(1)}_{t+1}-\alpha^{(1)}_t$ and $u_t^{(2)} = \alpha^{(2)}_{t+1}-\alpha^{(2)}_t$. The objective of the experiment is to learn how to control the motor angles in order to stabilize the ball around the center of the plate. The trial length is 3 seconds, with a sampling frequency of 30 [Hz]. The cost function encoding the task is
\begin{equation*}
   c(\boldsymbol{x}_t)=1-\text{exp}\left(-g(\boldsymbol{x}_t) \right), \qquad \text{with}
\end{equation*}
\begin{equation*}
    g(\boldsymbol{x}_t) = \left(\frac{b^x_t}{0.15}\right)^2 +\left(\frac{b^y_t}{0.15}\right)^2  +\left(\alpha_t^{(1)}\right)^2  +\left(\alpha_t^{(2)}\right)^2.
\end{equation*}

\begin{figure}[t]
\centering
  \includegraphics[width=0.95\linewidth]{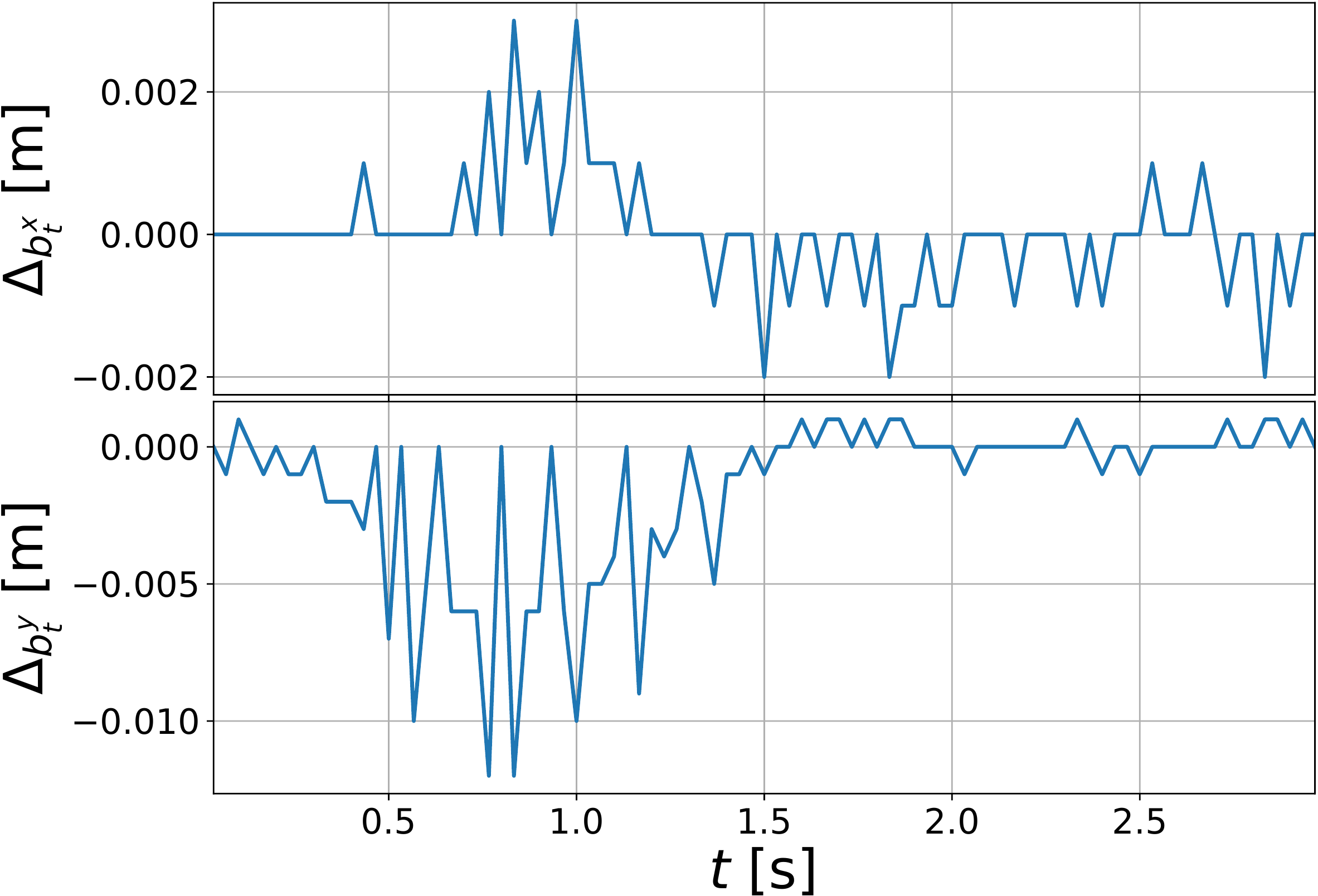}
  \caption{\small Example of GP targets in the ball-and-plate experiment, i.e. measured ball position changes in X and Y directions.}
  \label{fig:bpGPout}
  \vspace{-3mm}
\end{figure}

With regards to the VF model setup, we considered position memory $m_{\boldsymbol{q}}=2$ and control memory $m_{\boldsymbol{u}}=1$, and we replaced in both GP inputs $\tilde{\boldsymbol{x}}_{t}$ and policy input $\boldsymbol{x}^*$, the occurrences of $\alpha^{(1)}_t$ and $\alpha^{(2)}_t$ with their \textit{sin-cos} expansion. Analogously to the previous MC-PILCO4PMS experiment, the kernel function of the VF GP model is given by the sum of a SE kernel that takes as input the whole GP input vector, and of a linear kernel that takes as input only the \textit{sin-cos} expansion of angular quantities. The control policy is a \textit{squashed-RBF-network} with $n_b=400$ basis functions. Policy optimization involves the use of $M=400$ particles.

The initial exploration is implemented in two trials, in which the control signals are two distinct noisy triangular waves. Mostly during exploration and initial trials, the ball might touch the borders of the plate. In those cases, we kept data up to the collision instant and discarded it thereafter. The presence of quantization errors was simulated during particles generation by corrupting predicted angles with a uniform fictitious measurement noise $\mathcal{U}(\frac{-0.001}{2},\frac{0.001}{2})$ [m]. A peculiarity of this experiment in comparison to the others seen before is a wide range of initial conditions. In fact, the policy must learn how to control the ball to the center starting from any position on the plate's surface. Hence, the initial distribution considered for $b^x_0$ and $b^y_0$ is the uniform $\mathcal{U}(-0.15,0.15)$ [m]. 
\begin{figure}[t]
\centering
  \includegraphics[width=0.95\linewidth]{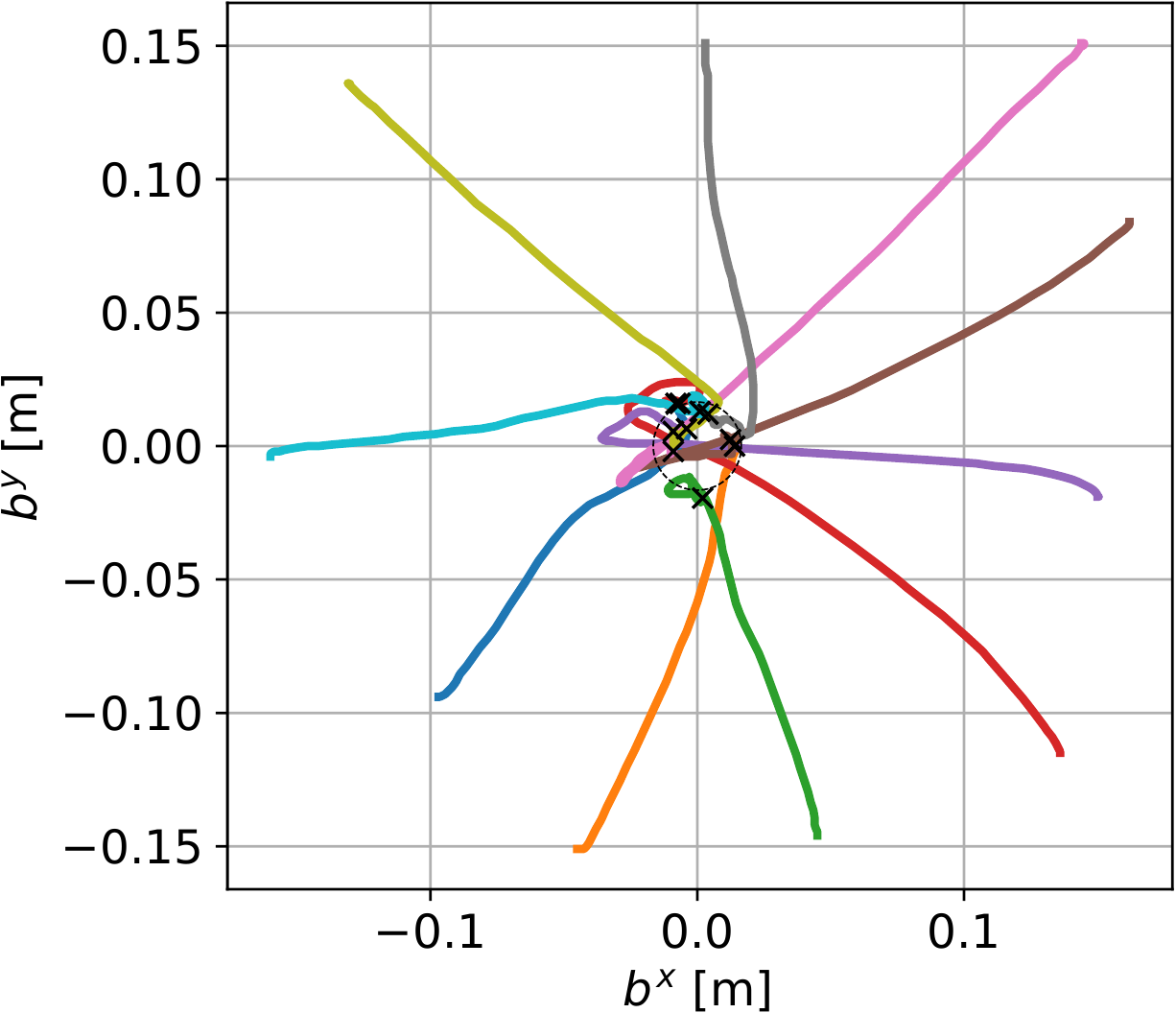}
  \caption{\small Ten different ball trajectories obtained by VF-MC-PILCO policy. Steady-state positions are marked with black crosses. The dashed circle has the same diameter as that of the ball.}
  \label{fig:b&p}
  \vspace{-3mm}
\end{figure}
The measurements provided by the camera are affected by a significant quantization error.
For instance, Fig. \ref{fig:bpGPout} reports the measured differences between consecutive ball positions during a trial. Consider that these quantities are the targets of the GPs in the VF model. 
In such a context,  ball velocity estimation can be very challenging.
In fact, for applying MC-PILCO4PMS on the same system, methods like finite differences and low-pass filtering were not sufficient, and it was necessary to implement a Kalman filter (online) and a Kalman smoother (offline), whose tuning was a delicate and time-consuming procedure of critical importance for the success of the algorithm.
On the contrary, VF-MC-PILCO managed to solve the task by working directly with raw position measurements, without the need of applying any kind of filtering.
Besides that, VF-MC-PILCO proved to be surprisingly data-efficient, being able to solve the task at the second trial, after only 7.97 seconds of experience, whereas MC-PILCO4PMS solved the task after 11.33 seconds. 

We tested the policy starting from ten different points in order to compare the two policies obtained by VF-MC-PILCO (Fig. \ref{fig:b&p}) and MC-PILCO4PMS. The mean steady-state error, i.e. the average distance of the last ball position from the center observed in the ten tests, was 0.0134 [m], while MC-PILCO4PMS final policy obtained a slightly better result, with a mean error of 0.0099 [m]. 
This might be due to the difference between the two policy inputs: MC-PILCO4PM relies on a Kalman filter, while VF-MC-PILCO works directly with raw measurements (in presence of significant noise).
Nevertheless, this performance difference is quite negligible, given the dimension of the ball whose radius is 0.016 [m].
% Nevertheless, this difference in terms of performance is quite negligible, given the dimension of the ball whose radius is 0.016 [m], and the fact that VF-MC-PILCO worked directly on raw measurements.

\section{Conclusions}\label{sec:conclusions}
We presented VF-MC-PILCO, a novel MBRL algorithm, specifically designed to learn from scratch how to control mechanical systems, without the need of computing any explicit velocity estimates. In our opinion, this may be a critical advantage when dealing with real systems affected by significant measurement noise, since, in this kind of scenario, the design of accurate velocity estimators can be a tedious task. The algorithm uses GPR to model the joint position changes, based on the history of past control actions and measurements.  VF-MC-PILCO was tested both in simulated environments (cart-pole and UR5 robot) as well as in two real mechanical systems (Furuta pendulum and ball-and-plate rig). It proved able to solve all the tasks, with results that are in line with the performance of our previous MBRL algorithm (MC-PILCO4PMS), which instead works with a complete state representation and must perform velocity estimation.

\bibliographystyle{unsrt}
\bibliography{references}

\end{document}